%% file: main.tex
\documentclass[11pt]{article}

\usepackage{acl}

\usepackage[linesnumbered, ruled, vlined]{algorithm2e}
\SetAlFnt{\small}

\usepackage{amsmath}
\usepackage{amssymb}
\usepackage[most]{tcolorbox}
\usepackage{tikz}
\usepackage{adjustbox}
\tcbuselibrary{skins, breakable}
\usepackage{tabularx}
\usepackage{multirow}
\usepackage{booktabs}
\usepackage{makecell}
\usepackage{colortbl}
\usepackage{ulem}
\usepackage{balance}
\usepackage{flushend}

\usepackage{times}
\usepackage{latexsym}

\usepackage[T1]{fontenc}
\usepackage{microtype}

\usepackage[utf8]{inputenc}

\usepackage{microtype}

\usepackage{inconsolata}
\usepackage{float}  

\usepackage{graphicx}
\usepackage{subcaption}
\usepackage{pgfplots}
\usetikzlibrary{patterns}
\usepgfplotslibrary{statistics}
\usepgfplotslibrary{groupplots}
\usepgfplotslibrary{fillbetween}
\usepackage{bbm}

\usepackage[utf8]{inputenc}
\usepackage{xcolor}
\usepackage{listings}
\usepackage{tcolorbox}
\usepackage{enumitem}
\tcbuselibrary{skins, breakable, listings}

\newtcolorbox{promptbox}{
  breakable,
  colback=gray!3,
  colframe=black,
  boxrule=0.4pt,
  arc=1pt,
  left=2mm,
  right=2mm,
  top=1mm,
  bottom=1mm,
  width=\linewidth,
  fontupper=\ttfamily\small,
  verbatim
}

\title{The Path Not Taken: Duality in Reasoning about Program Execution}

\author{
 \textbf{Eshgin Hasanov}, 
 \textbf{Md Mahadi Hassan Sibat}, 
 \textbf{Santu Karmaker}, 
 \textbf{Aashish Yadavally}
\\
 Department of Computer Science, University of Central Florida
\\
 \small{
   \textbf{Correspondence:} \href{mailto:eshgin.hasanov@ucf.edu}{eshgin.hasanov@ucf.edu}, \href{mailto:aashish.yadavally@ucf.edu}{aashish.yadavally@ucf.edu}
 }
}

\definecolor{violinA}{HTML}{D3D9AB}
\definecolor{violinB}{HTML}{E6BBD1}
\definecolor{violinC}{HTML}{B3D1B1}
\definecolor{gray}{rgb}{0.5,0.5,0.5}

\newcommand{\tool}{\textsc{DexBench}\xspace}

\definecolor{dkgreen}{rgb}{0,0.4,0}

\lstdefinestyle{exstyle}{
    basicstyle=\ttfamily\small,
    keywordstyle=\color{blue},
    commentstyle=\color{gray},
    stringstyle=\color{purple},
    breaklines=true,
    showstringspaces=false,
    columns=fullflexible
}

\begin{document}
\maketitle
\begin{abstract}

\input{sections/abstract}


\end{abstract}

\section{Introduction}
\input{sections/intro}

\section{Duality of Reasoning}\label{sec:duality}
\input{sections/duality}

\section{\tool}
\input{sections/methodology}

\section{Empirical Results}\label{sec:results}
\input{sections/discussion}

\section{Related Work}
\input{sections/related-short}

\section{Conclusion}
\input{sections/conclusion}

\section{Limitations}
\input{sections/limitation}


\newpage

\balance 

\bibliographystyle{latex/acl_natbib}
\bibliography{latex/custom}

\newpage
\appendix
\input{sections/appendix-final}

\end{document}

%% file: sections/abstract.tex
Large language models (LLMs) have shown remarkable capabilities across diverse coding tasks. However, their adoption requires a true understanding of program execution rather than relying on surface-level patterns. Existing benchmarks primarily focus on predicting program properties tied to specific inputs (\textit{e.g.}, code coverage, program outputs). As a result, they provide a narrow view of dynamic code reasoning and are prone to data contamination. We argue that understanding program execution requires evaluating its inherent duality through two complementary reasoning tasks: (\textit{i}) predicting a program's observed behavior for a given input, and (\textit{ii}) inferring how the input must be mutated toward a specific behavioral objective. 
Both tasks jointly probe a model's causal understanding of execution flow.
We instantiate this duality in \tool, a benchmark comprising 445 paired instances, and evaluate 13 LLMs.
Our results demonstrate that dual-path reasoning provides a robust and discriminative proxy for dynamic code understanding.

%% file: sections/intro.tex
Large language models (LLMs) are increasingly being adopted in the field of software engineering (SE), from writing code and unit tests to finding and fixing bugs. However, their performance remains inconsistent: these models can solve complex coding problems~\cite{DBLP:conf/iclr/JimenezYWYPPN24}, yet still fail to reason about basic loops~\cite{DBLP:journals/pacmse/DhulipalaYPN25}.
This can possibly be due to mimicking surface-level patterns without a \textit{true} understanding of the program, which inherently limits their applicability in real-world settings. 
As a result, recent efforts have focused on a fine-grained evaluation of the reasoning capabilities of coding LLMs~\cite{DBLP:journals/corr/abs-2402-09664,DBLP:conf/icse/ChenP000025,DBLP:journals/corr/abs-2510-15079}, moving beyond end-to-end evaluations on {downstream} tasks such as program synthesis~\cite{DBLP:journals/corr/abs-2107-03374,DBLP:conf/icml/GuRLSS024} or repair~\cite{DBLP:conf/iclr/JimenezYWYPPN24}.

In practice, such reasoning evaluations primarily rely on \textit{executability}, a distinct feature of source code requiring accurate modeling of execution flow. These assess how well LLMs understand program behaviors during execution, including predicting code coverage~\cite{DBLP:journals/pacmse/DhulipalaYPN25,DBLP:conf/icse/ChenP000025}, input-output mappings~\cite{DBLP:conf/nips/DingPMKYR24,DBLP:conf/icse/ChenP000025}, as well as tracking intermediate 
updates to program state and subsequent execution paths~\cite{DBLP:conf/acl/LiuLCJSFSD23,DBLP:conf/icse/ChenP000025}.

Nevertheless, current program execution-based benchmarks consider runtime behaviors under only a single test case, \textit{i.e.}, along \textit{one} execution
path (\textit{e.g.}, predicting the statements $\mathcal{S}$ in a program $P$ that are executed for an input $I$). However, a given program may traverse many different paths depending on its inputs. Therefore, single-test evaluations offer a narrow view of the models' program understanding. Moreover, since benchmarks defined by fixed sets of input-output pairs can be memorized during training~\cite{DBLP:conf/acl/RiddellNC24}, they are susceptible to data contamination, a growing concern in modern LLM evaluation.


To address this gap, we introduce the concept of \textbf{duality in reasoning about program execution}. The key idea is based on the argument that two program paths share a common execution space and diverge only at branching points where the current program state determines diverging execution flows. We refer to the actual path taken for a given input as \textit{execution path} and the alternative as \textit{counterfactual path}.
We operationalize our framework by designing a pair of complementary reasoning tasks: one predicting a program property observed along the execution path (\textit{forward reasoning}); other determining the input required for counterfactual path to be taken (\textit{backward reasoning}).

Note that a program naturally exhibits 
$n$-ality (or ``multiality''), 
characterized by many possible pairs of execution and counterfactual paths across its input domain. This property can be viewed as a composition of discrete dual pairs. By selecting one such pair for evaluation, we can isolate and evaluate the causal logic a model employs to reason about program execution.
Accordingly, we posit that probing a model's ability to understand both \textit{how} the observed execution occurs and under \textit{what} conditions (\textit{e.g.}, intermediate program state) the execution would instead flow through the counterfactual path provides a more robust evaluation of reasoning {capability} in coding LLMs.

\begin{figure}[t]
    \centering
    \includegraphics[width=0.95\linewidth]{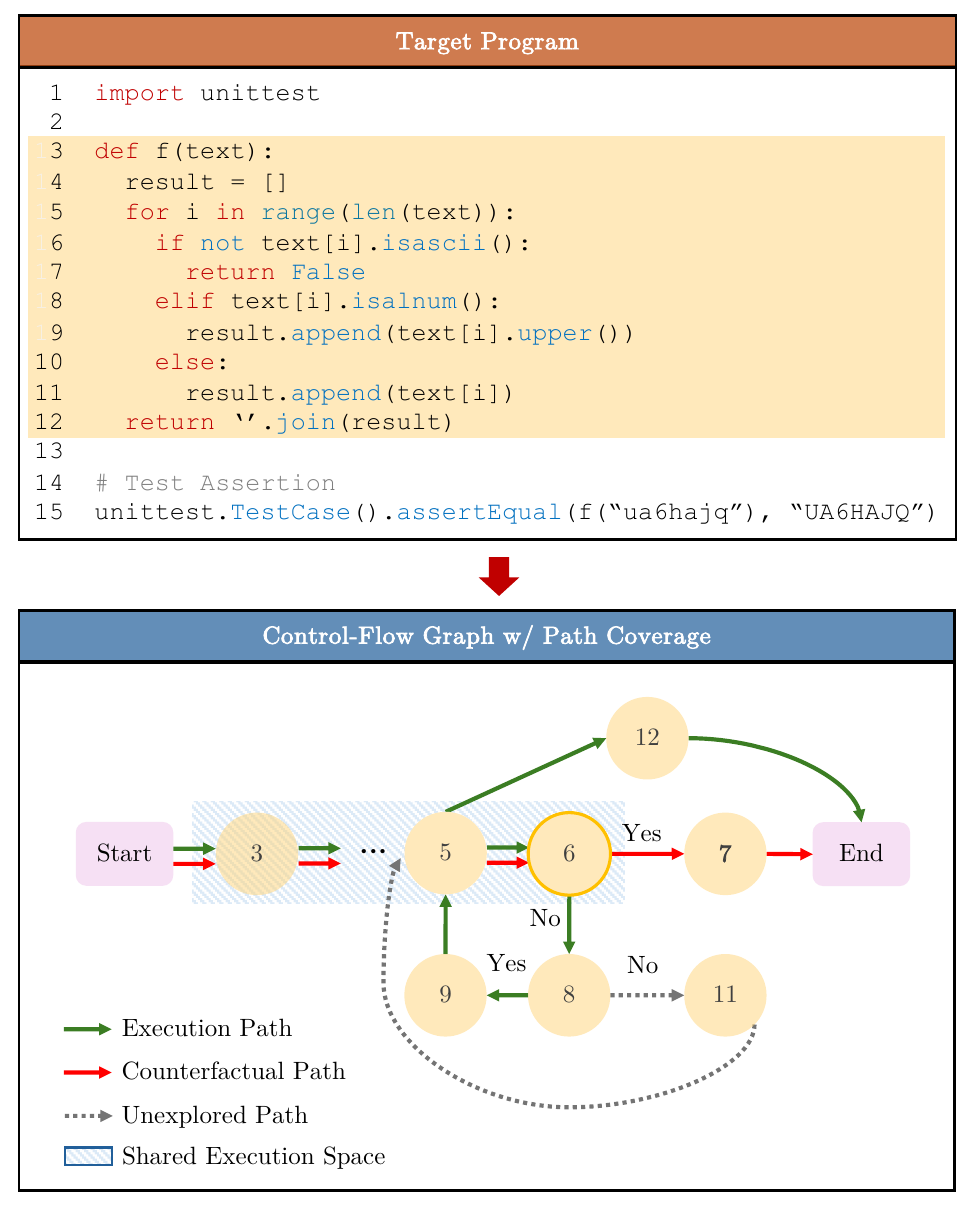}
    \caption{\tool exploits the shared execution space between execution and counterfactual paths to enable a robust evaluation of LLMs in program execution.}
    \vspace{-5pt}
    \label{fig:overview}
\end{figure}

As an illustration, consider the target program in Figure~\ref{fig:overview}. The program accepts multiple paths, including  $\pi^{(1)}\!:$ {\footnotesize\(3\!\to\!...\!\to\!6\!\to\!8\!\to9\to\!...\!\to\!6\!\to8\to\!9\!\to5\!\to\!12\!\to\!\mathbf{[END]}\)}, 
$\pi^{(2)}\!:$ {\footnotesize\(3\!\to\!...\!\to\!5\!\to\!6\!\to\!7\!\to\!\mathbf{[END]}\)}, $\pi^{(3)}\!:$ {\footnotesize\(3\!\to\!...\!\to\!6\!\to\!8\!\to\!...\!\to\!11\!\to\!5\!\to\!...\!\to\!\mathbf{[END]}\)}, among others. Any pair $\langle\pi^{(i)}, \pi^{(j)}\rangle$ reflects this multiality. For the test input ``{\footnotesize\texttt{ua6hajq}}'', execution follows the control-flow path $\pi^{(1)}$. Here, we designate $\pi^{(1)}$ as the \textit{execution path} $\pi_{exec}$ and select $\pi^{(2)}$ as the \textit{counterfactual path} $\pi_{cf}$ to illustrate the proposed duality (highlighted in Figure~\ref{fig:overview} in green and red, respectively)\footnote{In Section~\ref{sec:datasets}, we describe the algorithm used to select the counterfactual path, and in Section~\ref{sec:sensitivity}, we benchmark LLMs with different counterfactual path selection strategies.}. 
In particular, both $\pi_{exec}$ and $\pi_{cf}$ share execution space {\footnotesize\(3\!\to\!...\!\to\!6\)}, diverging only at line~6 depending on the branch condition. 
When it evaluates to ``{\texttt{No}}'', the execution proceeds along $\pi_{exec}$; otherwise, $\pi_{cf}$ is taken.

By design, forward reasoning evaluates an LLM's ability to predict program properties along $\pi_{exec}$ for a given input $I_{exec}$, while backward reasoning evaluates whether it can infer the new input $I_{cf}$ that $I_{exec}$ should be mutated to such that execution now flows along $\pi_{cf}$ instead. 
Notably, the novelty in the notion of duality lies in its focus on the causal relationships in program execution that induce divergence from $\pi_{exec}$ to $\pi_{cf}$, rather than in reasoning about program properties along distinct paths in isolation~\cite{DBLP:journals/corr/abs-2510-15079}.

Accordingly, we propose \tool, a \textbf{dual-path, program execution-based benchmark} that jointly investigates both forward execution reasoning and backward counterfactual reasoning. \tool is built from real-world programs that cover a range of complexity, extracted from widely used HumanEval~\cite{DBLP:journals/corr/abs-2107-03374}, CruxEval~\cite{DBLP:conf/icml/GuRLSS024}, and PythonSaga~\cite{DBLP:conf/emnlp/YadavB024}. We evaluate 9 open-source and 4 closed-source/proprietary LLMs, spanning both reasoning and general-purpose models across small- to large-scale model sizes. Our experiments reveal that:

(\textit{i}) strong performance on execution or counterfactual reasoning in isolation does not translate to success under joint evaluation, highlighting the limitations of single-path evaluation benchmarks.

(\textit{ii}) dual-path reasoning provides a reliable proxy for causal, state-aware reasoning of execution flow.

(\textit{iii}) parameter scaling and reasoning-focused post-training do not guarantee improvements in dual-path reasoning, as seen with some mid-sized and general-purpose models that outperformed larger or explicitly reasoning-focused counterparts.

Together, our findings indicate that \tool offers a systematic and extensible framework to evaluate dynamic code understanding in LLMs. 
All code and data are publicly available.\footnote{\url{https://github.com/sail-ucf/dexbench}}


%% file: sections/duality.tex
We formalize the concept of duality in reasoning about program execution as follows.
\subsection{(Forward) Execution Reasoning}\label{sec:duality-forward}

Consider a program $P$ with test input $I_{exec} \in \mathcal{I}$. Executing $P$ on $I_{exec}$ induces an execution trace 
\[
\tau_{exec}\!=\!\big(\langle s^{(1)}, \sigma^{(1)} \rangle ...\langle s^{(|\tau_{exec}|)}, \sigma^{(|\tau_{exec}|)} \rangle \big) \in \mathcal{T},
\]
where each $s^{(t)}$ is a statement in $P$ executed at step $t$, and $\sigma^{(t)}$ is the corresponding program state, defined as a mapping from program variables to their values. 
Let $\phi:\!\mathcal{T}\!\to\!\mathcal{O}$ denotes the observable program properties or runtime behaviors derived from execution traces (\textit{e.g.}, program output, code coverage, or a post-condition over variables). 

For program $P$ and input $I_{exec}$, we refer to the task of predicting an observable behavior $\phi(\tau_{exec})$ as \textit{forward or execution reasoning}. Formally, 
\begin{equation}\label{eq:forward}
\mathcal{R}_{exec}: (\textcolor{black}{P}, \textcolor{black}{I_{exec}}) \mapsto \textcolor{black}{\phi(\tau_{exec})}.
\end{equation}

By design, this formulation captures a model's ability to maintain a representation of the program state $\sigma^{(i)}$ and to update it according to the semantics of each executed statement $s^{(i)}$. The difficulty of execution reasoning typically scales with the length of the trace $|\tau_{exec}|$ as well as with the complexity of state transitions, such as those induced by non-trivial control flow~\cite{DBLP:conf/acl/LiuLCJSFSD23}.


\subsection{(Backward) Counterfactual Reasoning}\label{sec:duality-backward}

Let \textit{counterfactual objective} $\phi^\star \in \mathcal{O}$ be a \textit{desired} observable behavior that is obtained by applying a transformation $f\!:\!\mathcal{O}\!\to\!\mathcal{O}$ to the original behavior induced by $I_{exec}$ such that $\phi^\star=f\!\left(\textcolor{black}{\phi(\tau_{exec})}\right)$.


For program $P$, original input $I_{exec}$ and a given counterfactual objective $\phi^\star$, we define \textit{backward or counterfactual reasoning} as the task of mutating $I_{exec}$ into an alternative input $I_{cf} \in \mathcal{I}$ such that executing $P$ on $I_{cf}$ produces an execution trace $\tau_{cf}$ satisfying $\phi(\tau_{cf}) = \phi^\star$.
Formally,
\begin{equation}\label{eq:backward}
\mathcal{R}_{cf} : (\textcolor{black}{P}, \textcolor{black}{I_{exec}}, \phi^\star) \mapsto I_{cf}.    
\end{equation}

This formulation captures the inverse capability of a model, \textit{i.e.}, to reason backwards from a desired effect to the conditions that would produce it. In contrast to execution reasoning, which follows the program’s semantics forward from a given input, counterfactual reasoning requires analyzing how changes to the original input affect control flow and program state in order to realize a behavioral goal (\textit{e.g.}, execution of an otherwise uncovered branch, specific program output). The difficulty of counterfactual reasoning typically increases with the complexity of the program’s branching structure and the degree to which the counterfactual execution diverges from the original one.

\subsection{Dual-Path Reasoning}
We define \textit{dual-path reasoning} $\mathcal{R}_{dual}$ as the task of jointly reasoning over the original execution path $\pi_{exec}$ and a counterfactual path $\pi_{cf}$. Formally,
\begin{equation}\label{eq:dual}
\mathcal{R}_{dual}: \mathcal{R}_{exec} \oplus \mathcal{R}_{cf} 
\end{equation}
where ``$\oplus$'' denotes the composition of execution and counterfactual reasoning. 
It probes a model's ability to: (\textit{i}) maintain an observable behavior produced along $\pi_{exec}$, and (\textit{ii}) manipulate runtime behavior to identify an alternative input $I_{cf}$ that satisfies a counterfactual objective $\phi^\star$, defined relative to the original execution. Such a unified perspective is particularly important for SE tasks that require a systematic exploration of program behavior (\textit{e.g.}, coverage-guided fuzzing, debugging, \textit{etc.}), where understanding how and why executions diverge is as important as predicting their outcomes. 
Furthermore, dual-path reasoning enables a robust evaluation of coding LLMs as it demands a deeper, more holistic understanding of program execution than reasoning on single, isolated paths.

%% file: sections/methodology.tex
\subsection{Problem Definition}

To evaluate the dual-path reasoning capabilities of LLMs, we instantiate the formulations for execution and counterfactual reasoning (Eqs.~\ref{eq:forward} and~\ref{eq:backward}) as concrete, verifiable tasks: \textit{code coverage prediction} and \textit{branch-targeted input mutation}, respectively.
For execution reasoning, the observable behavior that a model must predict is \textit{statement coverage}, \textit{i.e.}, the {set of line numbers} $\phi(\tau_{exec})$ executed by a program under input $I_{exec}$. 
In the case of the latter, we define the counterfactual objective $\phi^\star$ as the \textit{reachability of a specific branch $b$ that remains uncovered under $I_{exec}$}. Here, the model must generate a mutated input $I_{cf}$ that redirects the program's execution to flow through $b$ such that $b \in \phi(\tau_{cf})$.

Unlike coverage-guided fuzzing~\cite{DBLP:conf/ccs/BohmePNR17}, which relies on random mutations and execution feedback, \tool requires models to identify necessary input changes through reasoning alone. By grounding both execution and counterfactual prediction in a shared execution context, we evaluate whether models maintain a consistent causal representation of program behavior. This prevents the evaluation from treating execution and generation as decoupled tasks, instead probing the model's ability to logically bridge observed runtime behaviors with desired behavioral outcomes. 

Note that our proposed framework can also be extended to other evaluation settings. For instance, execution reasoning may target predicting program outputs or intermediate states, while counterfactual reasoning may require predicting inputs that trigger specific program outputs or state transitions.



\subsection{Benchmark Construction}\label{sec:datasets}


\paragraph{Source Programs.}
\tool is constructed from a diverse collection of Python programs from three popular datasets: \textit{CruxEval}~\cite{DBLP:conf/icml/GuRLSS024}, \textit{HumanEval}~\cite{DBLP:journals/corr/abs-2107-03374}, and \textit{PythonSaga}~\cite{DBLP:conf/emnlp/YadavB024}. These are designed to evaluate program understanding and execution reasoning, making them well-suited for 
input-dependent reasoning over program executions. 

CruxEval contains short programs with relatively simple control flow. In contrast, HumanEval has moderate control-flow complexity, while PythonSaga includes programs with deeper nesting, recursion, and more advanced programming constructs. 
In Figure~\ref{fig:dataset-complexity}, we illustrate the program complexity for each dataset with different structural metrics. Based on this, we only retained programs that contain: (\textit{i}) at least one conditional or loop, and (\textit{ii}) a test case that yields partial code coverage (\textit{i.e.}, less than 100\%). The first criterion ensures reasoning over non-trivial control flow, while the second enables coverage-guided input mutation as required in the counterfactual or backward reasoning task.


\vspace{-2.5pt}
\paragraph{Collecting Execution Information.}
For each retained program, we randomly selected a test case meeting the above criteria to instantiate the execution or forward reasoning task, collecting statement coverage information 
by monitoring runtime behavior with Slipcover~\cite{slipcover}.
Here, coverage information serves as \textit{ground-truth observable behavior} (denoted by $\phi$ in Section~\ref{sec:duality-forward}).

\vspace{-2.5pt}
\paragraph{Selecting Counterfactual Paths.}
As illustrated in Figure~\ref{fig:overview}, the execution of a program under input $I_{exec}$ flows along a single path, leaving multiple alternative paths unexplored that could serve as potential counterfactual targets.
Among these, we select an uncovered branch $b$ under input $I_{exec}$ that, if reached, would enhance the execution diversity between the original execution path $\pi_{exec}$ and the counterfactual path $\pi_{cf}$ while resulting in an increased code coverage (see Appendix~\ref{app:line-priority} for detailed algorithm). Here, reaching branch $b$ thus serves as the counterfactual objective (denoted by $\phi^\star$ in Section~\ref{sec:duality-backward}). In this setting, the \textit{ground-truth counterfactual observable behavior} is defined by the program's execution semantics: a model-generated input $I_{cf}$ is considered correct if and only if its induced execution trace $\tau_{cf}$ satisfies $b \in \phi(\tau_{cf})$.

\vspace{-2.5pt}
\paragraph{Benchmark Statistics.} 
In total, \tool comprises 445 unique evaluation instances, each defined by a pair of forward and backward reasoning tasks (298 from CruxEval, 100 from HumanEval, and 47 from PythonSaga). It exhibits significant structural diversity: subject programs have 10 to 78 lines of code 
and a mean cyclomatic complexity of $3.63\!\pm\!{2.18}$ (ranging between 2 and 19),
thus yielding a challenging testbed for evaluating models’ ability to reason about program execution.

\input{diagrams/violin_grid}
\input{tables/intrinsic2}

\subsection{Baseline Models}
We select a diverse set of open-source LLMs, spanning: reasoning and non-reasoning variants; small- (<10B), mid- (20B--30B), and large-size (30B--70B) models. In particular, we include Jamba Reasoning-3B~\cite{DBLP:conf/iclr/LenzLABMPAAFPGJ25}, Llama-3.1 Nemotron Nano~\cite{DBLP:journals/corr/abs-2505-00949}, Llama-3.2-3B-Instruct and Llama-3.3-70B-Instruct~\cite{DBLP:journals/corr/abs-2407-21783}, Mistral Small-24B~\cite{DBLP:journals/corr/abs-2310-06825}, Magistral Small~\cite{DBLP:journals/corr/abs-2506-10910}, as well as QwQ-32B, Qwen2.5-32B and Qwen2.5-72B~\cite{DBLP:journals/corr/abs-2505-09388}.
We also benchmark against closed-source or frontier models, including Gemini 2.5 Flash~\cite{gemini25}, GPT-5 Mini~\cite{gpt5_system_card}, Claude Sonnet 4~\cite{claude_sonnet4}, and Grok-4 Reasoning~\cite{grok4}.

\subsection{Prompts}
We use two dedicated prompt templates to evaluate execution (forward) and counterfactual (backward) reasoning, respectively. To ensure fair comparison, we apply the same prompt structure and formatting constraints across all models. We adopt one-shot prompting (\textit{i.e.}, with a single illustrative example) for all experiments on the three benchmarks. We also analyze robustness of \tool by probing the model sensitivity to more complex prompts (Section~\ref{sec:results}). See Appendix~\ref{sec:prompt-design} for all templates.

\subsection{Evaluation Metrics}\label{sec:eval-metrics}
For both execution (forward) and counterfactual (backward) reasoning, we evaluate performance using the $\mathrm{pass@}k$ metric~\cite{DBLP:journals/corr/abs-2107-03374}. In execution reasoning, an instance is considered successful if at least one of the $k$ generated candidates predicts code coverage exactly correctly. Formally, the success indicator $S_{exec}$ for the instance is
\begin{equation*} 
S_{exec}\!=\!\mathbbm{1}\!\left(\exists \tilde{\phi}_i\!\in\!\{\tilde{\phi}_1, ..., \tilde{\phi}_k\} \; \text{s.t.} \; \tilde{\phi}_i\!=\!\phi(\tau_{exec}) \right),
\end{equation*}
where $\tilde{\phi}_i$ denotes the code coverage predicted by the $i$-th candidate.

In counterfactual reasoning, an instance is considered successful if at least one of the $k$ generated test inputs covers the originally uncovered branch $b$ (as identified in Section~\ref{sec:datasets}). Formally, the success indicator $S_{cf}$ for the instance is defined as
\begin{equation*}
S_{cf}\!=\!\mathbbm{1} \left( \exists I_{cf, i}\!\in\!\{I_{cf, 1}, ..., I_{cf, k}\} \; \text{s.t.} \; b\!\in\!\phi(\tau_{cf, i}) \right),    
\end{equation*}
where $\tau_{cf,i}$ is the execution trace produced by running the corresponding program on input $I_{cf,i}$.

Putting them together, we define a joint success indicator $S_{dual}$ for dual-path reasoning as
\begin{equation*}
S_{dual} = S_{exec} \wedge S_{cf}.
\end{equation*}
In other words, an instance is successful under the joint metric only if the model both correctly predicts the original coverage and generates at least one input whose execution covers the target branch.

%% file: diagrams/violin_grid.tex
\begin{figure}[t]
    \centering
    \resizebox{0.85\linewidth}{!}{
    \begin{minipage}{\linewidth}
    \begin{tabular}{|rlrlrl|}
        \hline
        \textcolor{violinA}{\rule{0.75em}{0.5em}} & {\footnotesize CruxEval} &
        \textcolor{violinB}{\rule{0.75em}{0.5em}} & {\footnotesize HumanEval} &
        \textcolor{violinC}{\rule{0.75em}{0.5em}} & {\footnotesize PythonSaga} \\
        \hline
    \end{tabular}

    \vspace{0.1cm}

    \begin{subfigure}{0.485\linewidth}
        \centering
        \includegraphics[width=\linewidth]{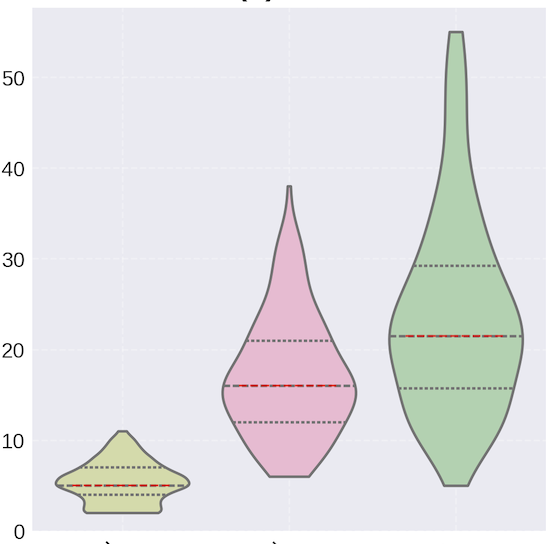}
        \caption{Lines of Code}
    \end{subfigure}
    \hspace{0.05cm}
    \begin{subfigure}{0.485\linewidth}
        \centering
        \includegraphics[width=\linewidth]{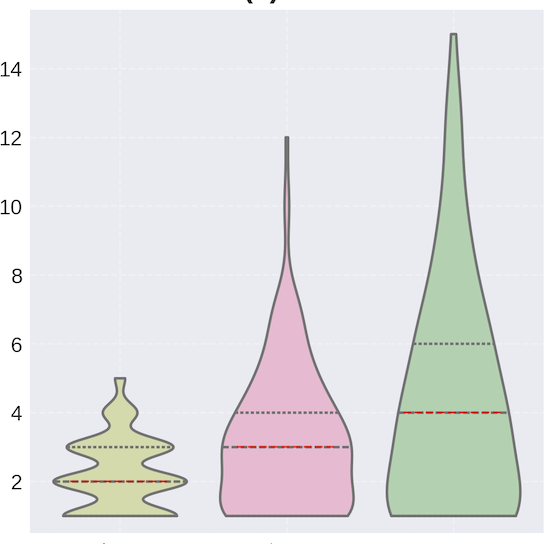}
        \caption{Cyclomatic Complexity}
    \end{subfigure}

    \vspace{0.1cm}

    \begin{subfigure}{0.485\linewidth}
        \centering
        \includegraphics[width=\linewidth]{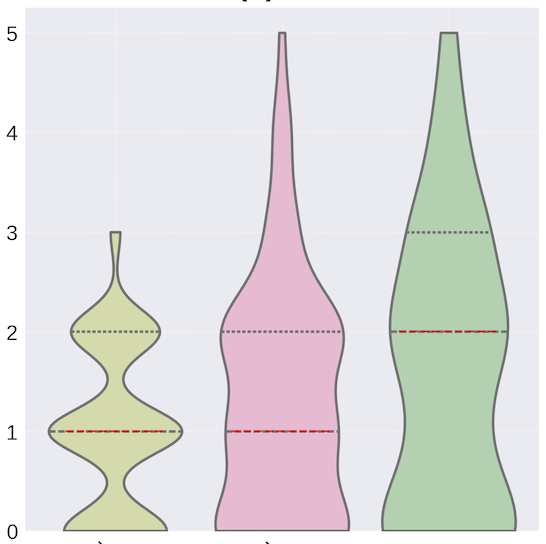}
        \caption{Nested Constructs}
    \end{subfigure}
    \hspace{0.05cm}
    \begin{subfigure}{0.485\linewidth}
        \centering
        \includegraphics[width=\linewidth]{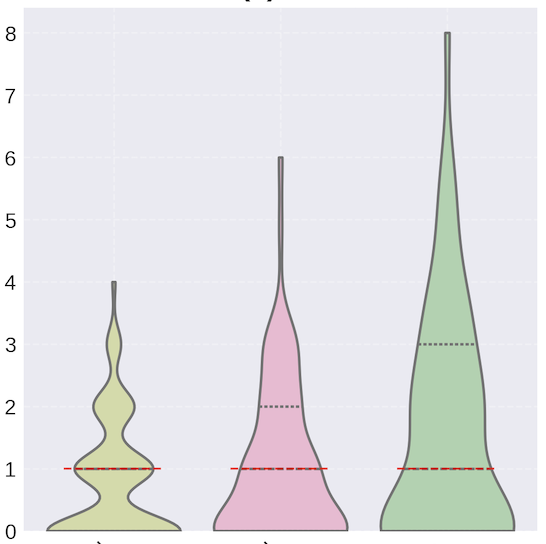}
        \caption{Loop Lengths}
    \end{subfigure}

    \end{minipage}
    }
    \caption{Distribution of different program complexity metrics in CruxEval, HumanEval, and PythonSaga.}
    \label{fig:dataset-complexity}
\end{figure}

%% file: tables/intrinsic2.tex
\begin{table*}[t]
\centering
\setlength{\tabcolsep}{4pt} 
\resizebox{\textwidth}{!}{
\begin{tabular}{c|cccccc|cccccc|cccccc}
\toprule
\multicolumn{1}{c|}{\multirow{3}{*}{\makecell{\textbf{Dataset ($\rightarrow$)} \\ \textbf{Model ($\downarrow$)}}}}                                & \multicolumn{6}{c|}{\textbf{CruxEval}}                                                       & \multicolumn{6}{c|}{\textbf{HumanEval}}                                                      & \multicolumn{6}{c}{\textbf{PythonSaga}}                                        \\ 
\cmidrule{2-19} 
\multicolumn{1}{c|}{}                                & \multicolumn{2}{c}{$\mathcal{R}_{exec}$} & \multicolumn{2}{c}{$\mathcal{R}_{cf}$} & \multicolumn{2}{c|}{$\mathcal{R}_{dual}$} & \multicolumn{2}{c}{$\mathcal{R}_{exec}$} & \multicolumn{2}{c}{$\mathcal{R}_{cf}$} & \multicolumn{2}{c|}{$\mathcal{R}_{dual}$} & \multicolumn{2}{c}{$\mathcal{R}_{exec}$} & \multicolumn{2}{c}{$\mathcal{R}_{cf}$} & \multicolumn{2}{c}{$\mathcal{R}_{dual}$} \\ 
\multicolumn{1}{c|}{}                                & $k\!=\!1$ & $k\!=\!5$ & $k\!=\!1$ & $k\!=\!5$ & $k\!=\!1$ & $k\!=\!5$ & $k\!=\!1$ & $k\!=\!5$ & $k\!=\!1$ & $k\!=\!5$ & $k\!=\!1$ & $k\!=\!5$ & $k\!=\!1$ & $k\!=\!5$ & $k\!=\!1$ & $k\!=\!5$ & $k\!=\!1$ & $k\!=\!5$ \\
\midrule
\multicolumn{1}{c|}{Jamba Reasoning 3B} & 0.0 & {\cellcolor{gray!12}{0.0}} & 9.4 & {\cellcolor{gray!12}{14.8}} & 0.0 & {\cellcolor{gray!12}{0.0}} & 0.0 & {\cellcolor{gray!12}{0.0}} & 15.0 & {\cellcolor{gray!12}{31.0}} & 0.0 & {\cellcolor{gray!12}{0.0}} & 0.0 & {\cellcolor{gray!12}{0.0}} & 10.6 & {\cellcolor{gray!12}{12.8}} & 0.0 & {\cellcolor{gray!12}{0.0}} \\
\multicolumn{1}{c|}{Nemotron Nano} & 3.0 & {\cellcolor{gray!12}{8.1}} & 15.8 & {\cellcolor{gray!12}{30.9}} & 0.3 & {\cellcolor{gray!12}{2.7}} & 0.0 & {\cellcolor{gray!12}{0.0}} & 15.0 & {\cellcolor{gray!12}{30.0}} & 0.0 & {\cellcolor{gray!12}{0.0}} & 0.0 & {\cellcolor{gray!12}{0.0}} & 10.6 & {\cellcolor{gray!12}{34.0}} & 0.0 & {\cellcolor{gray!12}{0.0}} \\
\multicolumn{1}{c|}{Llama-3.2-3B-Inst.} & 0.0 & {\cellcolor{gray!12}{0.0}} & 12.8 & {\cellcolor{gray!12}{32.6}} & 0.0 & {\cellcolor{gray!12}{0.0}} & 0.0 & {\cellcolor{gray!12}{0.0}} & 21.0 & {\cellcolor{gray!12}{43.0}} & 0.0 & {\cellcolor{gray!12}{0.0}} & 0.0 & {\cellcolor{gray!12}{0.0}} & 4.3 & {\cellcolor{gray!12}{27.7}} & 0.0 & {\cellcolor{gray!12}{0.0}} \\
\midrule
\multicolumn{1}{c|}{Mistral Small 24B} & 29.2 & {\cellcolor{gray!12}{69.8}} & 43.6 & {\cellcolor{gray!12}{63.4}} & 13.4 & {\cellcolor{gray!12}{43.6}} & 10.0 & {\cellcolor{gray!12}{35.0}} & 50.0 & {\cellcolor{gray!12}{61.0}} & 4.0 & {\cellcolor{gray!12}{20.0}} & 6.4 & {\cellcolor{gray!12}{6.4}} & 44.7 & {\cellcolor{gray!12}{53.2}} & 2.1 & {\cellcolor{gray!12}{2.1}} \\
\multicolumn{1}{c|}{Magistral Small} & 27.2 & {\cellcolor{gray!12}{62.4}} & 18.5 & {\cellcolor{gray!12}{34.2}} & 3.0 & {\cellcolor{gray!12}{21.5}} & 13.0 & {\cellcolor{gray!12}{40.0}} & 21.0 & {\cellcolor{gray!12}{30.0}} & 1.0 & {\cellcolor{gray!12}{16.0}} & 4.3 & {\cellcolor{gray!12}{14.9}} & 23.4 & {\cellcolor{gray!12}{36.2}} & 2.1 & {\cellcolor{gray!12}{4.3}} \\
\multicolumn{1}{c|}{QwQ-32B} & 20.1 & {\cellcolor{gray!12}{53.4}} & 50.7 & {\cellcolor{gray!12}{63.1}} & 11.4 & {\cellcolor{gray!12}{33.2}} & 21.0 & {\cellcolor{gray!12}{38.0}} & 61.0 & {\cellcolor{gray!12}{76.0}} & 13.0 & {\cellcolor{gray!12}{25.0}} & 0.0 & {\cellcolor{gray!12}{2.1}} & 53.2 & {\cellcolor{gray!12}{80.9}} & 0.0 & {\cellcolor{gray!12}{2.1}} \\
\multicolumn{1}{c|}{Qwen2.5-32B} & 34.2 & {\cellcolor{gray!12}{71.1}} & 57.7 & {\cellcolor{gray!12}{79.5}} & 18.1 & {\cellcolor{gray!12}{57}} & 11.0 & {\cellcolor{gray!12}{37.0}} & 64.0 & {\cellcolor{gray!12}{77.0}} & 8.0 & {\cellcolor{gray!12}{31.0}} & 0.0 & {\cellcolor{gray!12}{8.5}} & 46.8 & {\cellcolor{gray!12}{55.3}} & 0.0 & {\cellcolor{gray!12}{6.4}} \\
\midrule
\multicolumn{1}{c|}{Llama-3.3-70B-Inst.} & 15.8 & {\cellcolor{gray!12}{38.6}} & 49.3 & {\cellcolor{gray!12}{68.1}} & 8.7 & {\cellcolor{gray!12}{27.5}} & 5.0 & {\cellcolor{gray!12}{16.0}} & 50 & {\cellcolor{gray!12}{64}} & 3.0 & {\cellcolor{gray!12}{9.0}} & 0.0 & {\cellcolor{gray!12}{0.0}} & 51.1 & {\cellcolor{gray!12}{61.7}} & 0.0 & {\cellcolor{gray!12}{0.0}} \\
\multicolumn{1}{c|}{Qwen2.5-72B} & 34.9 & {\cellcolor{gray!12}{74.8}} & 41.3 & {\cellcolor{gray!12}{52.3}} & 14.1 & {\cellcolor{gray!12}{39.3}} & 10.0 & {\cellcolor{gray!12}{31.0}} & 41.0 & {\cellcolor{gray!12}{50.0}} & 1.0 & {\cellcolor{gray!12}{12.0}} & 0.0 & {\cellcolor{gray!12}{0.0}} & 38.3 & {\cellcolor{gray!12}{63.8}} & 0.0 & {\cellcolor{gray!12}{0.0}} \\
\midrule
\multicolumn{1}{c|}{Gemini 2.5 Flash} & 67.4 & {\cellcolor{gray!12}{85.9}} & 72.1 & {\cellcolor{gray!12}{85.2}} & 51.0 & {\cellcolor{gray!12}{73.8}} & 50.0 & {\cellcolor{gray!12}{77.0}} & 69.0 & {\cellcolor{gray!12}{80.0}} & 41.0 & {\cellcolor{gray!12}{62.0}} & 6.4 & {\cellcolor{gray!12}{12.8}} & 57.4 & {\cellcolor{gray!12}{66.0}} & 4.3 & {\cellcolor{gray!12}{\textcolor{black}{8.5}}} \\
\multicolumn{1}{c|}{GPT-5 Mini} & 89.9 & {\cellcolor{gray!12}{95.6}} & 81.2 & {\cellcolor{gray!12}{97.0}} & 73.8 & {\cellcolor{gray!12}{91.6}} & 79.0 & {\cellcolor{gray!12}{88.0}} & 77.0 & {\cellcolor{gray!12}{88.0}} & 59.0 & {\cellcolor{gray!12}{76.0}} & 36.2 & {\cellcolor{gray!12}{46.8}} & 83.0 & {\cellcolor{gray!12}{95.7}} & 29.8 & {\cellcolor{gray!12}{44.7}} \\
\multicolumn{1}{c|}{Grok-4 Reasoning} & 85.6 & {\cellcolor{gray!12}{86.9}} & 93.0 & {\cellcolor{gray!12}{97.0}} & 79.9 & {\cellcolor{gray!12}{84.9}} & 78.0 & {\cellcolor{gray!12}{84.0}} & 85.0 & {\cellcolor{gray!12}{88.0}} & 65.0 & {\cellcolor{gray!12}{72.0}} & 55.3 & {\cellcolor{gray!12}{68.1}} & 93.6 & {\cellcolor{gray!12}{95.7}} & 51.1 & {\cellcolor{gray!12}{66.0}} \\
\multicolumn{1}{c|}{Claude Sonnet 4} & 91.9 & {\cellcolor{gray!12}{98.7}} & 94.3 & {\cellcolor{gray!12}{96.3}} & \textbf{87.6} & {\cellcolor{gray!12}{\textbf{95.3}}} & 83.0 & {\cellcolor{gray!12}{92.0}} & 84.0 & {\cellcolor{gray!12}{87.0}} & \textbf{69.0} & {\cellcolor{gray!12}{\textbf{79.0}}} & 59.6 & {\cellcolor{gray!12}{72.3}} & 91.5 & {\cellcolor{gray!12}{95.7}} & \textbf{55.3} & {\cellcolor{gray!12}{\textbf{70.2}}} \\

\bottomrule
\end{tabular}
}
\vspace{-2.5pt}
\caption{Model performance comparison on execution reasoning ($\mathcal{R}_{exec}$) and counterfactual reasoning ($\mathcal{R}_{cf}$) tasks, as well as our dual-path reasoning ($\mathcal{R}_{dual}$) framework across CruxEval, HumanEval, and PythonSaga datasets (ordered left-to-right by increasing program complexity). Results are reported using $\mathrm{pass@}k$ metrics (in \%).}
\vspace{-5pt}
\label{tab:leaderboard}
\end{table*}

%% file: sections/discussion.tex
\subsection{Performance Comparison in Reasoning about Program Execution}\label{sec:model-comparison}



In Table~\ref{tab:leaderboard}, we compare model performance in execution reasoning ($\mathcal{R}_{exec}$), counterfactual reasoning ($\mathcal{R}_{cf}$), as well as their dual-path formulation ($\mathcal{R}_{dual}$). Our analysis considers three dimensions:

\textbf{(1) Program Complexity.} While closed-source or frontier models generally outperform the open-source ones, performance on dual-path reasoning consistently drops as program complexity increases (CruxEval $\!<\!$ HumanEval $\!<\!$ PythonSaga). On the PythonSaga dataset featuring intricate control-flow logic, Gemini 2.5 Flash records particularly low $\mathrm{pass@}1$ and $\mathrm{pass@}5$ scores of 4.3\% and 8.5\%, respectively. These results are comparable to those of the mid- and large-size open-source models and are significantly lower compared to other closed-source models, whose scores range from 29.8\%--55.3\% ($\mathrm{pass@}1$) and 44.7\%--70.2\% ($\mathrm{pass@}5$). 
Overall, increasing program complexity impacts success in joint reasoning about execution and counterfactual paths, as greater path divergence makes it difficult for models to maintain a consistent causal representation of the program.

\textbf{(2) Model Scaling Laws.} All small-scale models ($<\!$10B parameters) consistently fail in dual-path reasoning due to achieving near-zero performance on code coverage prediction (\textit{i.e.}, execution reasoning, $\mathcal{R}_{exec}$) despite non-trivial results in branch-guided input mutation (\textit{i.e.}, counterfactual reasoning, $\mathcal{R}_{cf}$). Scaling to mid-sized models (20B--30B parameters) yields substantial improvements in both reasoning tasks. However, these trends do not extend to large-size models ($>\!30$B parameters). In particular, within the Qwen family, the 32B parameter variant outperforms the 72B model in dual-path reasoning across all datasets. 


\textbf{(3) Reasoning-Focused Post Training.} Among the open-source models, we considered two pairs of medium-sized non-reasoning and reasoning variants: $\langle$Mistral Small 24B, Magistral Small$\rangle$, $\langle$Qwen2.5-32B, QwQ-32B$\rangle$. In both pairs, non-reasoning models surprisingly outperform the reasoning ones in dual-path reasoning by an average of 63.9\% and 47.9\% ($\mathrm{pass@}5$) on CruxEval and HumanEval. This trend persists even on the complex PythonSaga dataset, where Qwen2.5-32B (6.4\%) outperforms QwQ-32B (2.1\%), although Magistral Small (4.3\%) exhibits a marginal improvement over its non-reasoning counterpart (2.1\%). \textit{These results indicate that such post-training does not generalize to reasoning about program execution.}

In contrast, closed-source models demonstrate the best performance in dual-path reasoning across all datasets (Claude Sonnet 4 $\!>\!$ Grok-4 Reasoning $\!>\!$ GPT-5 Mini $\!>\!$ Gemini 2.5 Flash). However, due to a lack of transparency in their design, these gains cannot be directly attributed to reasoning-focused post-training alone and instead serve as reference upper bounds on achievable performance.


\subsection{Execution vs. Counterfactual Reasoning}\label{sec:direction}

In this section, we investigate the asymmetry in model performance (as in Section~\ref{sec:model-comparison}) in both forward and backward reasoning tasks, isolating \textit{where}, and \textit{to what extent} models break down.

\paragraph{Diagnosing Asymmetry in Execution and Counterfactual Reasoning.} 
In Table~\ref{tab:leaderboard}, we reported a consistent asymmetry in model performance on execution and counterfactual reasoning. Furthermore, as illustrated in Figure~\ref{fig:quadrant-stacked}, many models succeed for only one of the two tasks. Here, the execution-only setting (\textit{i.e.}, $\mathcal{R}_{exec}\!: \checkmark$, $\mathcal{R}_{cf}\!: \times$) reflects a model's ability to predict observed program behavior for a fixed input, while failing to reason about how changes to intermediate program state induce alternative execution flows. In contrast, the counterfactual-only setting (\textit{i.e.}, $\mathcal{R}_{exec}\!: \times$, $\mathcal{R}_{cf}\!: \checkmark$) suggests that a model can heuristically mutate inputs to alter execution flow without reliably predicting the execution behavior itself. When considered in isolation, both evaluation settings probe complementary model capabilities, as evidenced among the open-source models. 
However, a joint evaluation under dual-path reasoning offers a systematic framework for quantifying reasoning capabilities of LLMs in program execution, enabling rigorous and comparative analysis across models.

\vspace{-2.5pt}
\paragraph{Execution Reasoning under Relaxed Evaluation.} Figure~\ref{fig:strict-relaxed} reports the execution reasoning performance of all models across three datasets in two evaluation settings: \textit{strict} and \textit{relaxed}. The former corresponds to the binary success indicator defined in Section~\ref{sec:eval-metrics}, whereas the latter measures partial correctness using Jaccard similarity between the predicted ($\tilde{\phi}$) and ground-truth ($\phi(\tau_{exec})$) coverage sets. The relaxed success score for an instance is:
 \[
S_{exec}^{r} = \frac{|\tilde{\phi} \cap \phi(\tau_{exec})|}{|\tilde{\phi} \cup \phi(\tau_{exec})|}.
\]

\input{diagrams/direction2}

\input{diagrams/relaxed}

Here, a larger gap between strict and relaxed performance indicates that models often identify a substantial subset of the correct code coverage while failing the exact-match criteria, and the width of the shaded region
further captures variability in performance across datasets of increasing program complexity. Notably, the mid-size open-source reasoning model QwQ-32B exhibits less precise per-statement execution prediction than the small-size models as well. In contrast, larger open-source models achieve relaxed performance comparable to that of closed-source models like Gemini-2.5 Flash and GPT-5 Mini. Among all models, Grok-4 Reasoning and Claude Sonnet 4 make the least per-statement errors while exhibiting consistent performance across all three datasets, indicating their superior execution reasoning capabilities.

\input{diagrams/ablation}

\paragraph{Counterfactual Reasoning w/o Test Input.} In the counterfactual reasoning task, models mutate a given test input to satisfy specific branch coverage objectives. In this experiment, we remove the test input and instead require models to directly generate an input that reaches the target branch, \textit{i.e.}, without conditioning on the test input, thus transforming branch-targeted input mutation into branch-targeted input generation (see Appendix~\ref{sec:prompt-design}).

Table~\ref{tab:ablation} reports the performance of the well-performing GPT-5 Mini and Grok-4 Reasoning (from Section~\ref{sec:model-comparison}). 
Contrary to initial expectations, we observed better performance under input generation compared to the mutation setting, improving $\mathrm{pass@}1$ and $\mathrm{pass@}5$ scores across three datasets by 11.9\% and 4.4\% for GPT-5 Mini, and by 6.6\% and 4.9\% for Grok-4 Reasoning.
However, we soon realized that this is due to the relaxation of causal constraints in the input generation setting, where models no longer require to reason about how specific input perturbations propagate through intermediate program states. 
As such, these results 
support our original design choice by highlighting the greater difficulty of branch-targeted input mutation and thereby, motivating its use in \tool as the counterfactual reasoning task.

\subsection{Sensitivity Analysis}\label{sec:sensitivity}

In this section, we examine the robustness of model performance in both execution and counterfactual reasoning under different perturbations.

\input{diagrams/sensitivity-forward}

\paragraph{Robustness to Prompt Complexity.} 
To assess the robustness of execution reasoning within \tool under increased prompt complexity, we replace our prompt for predicting code coverage with that from the state-of-the-art CRISPE~\cite{DBLP:journals/pacmse/DhulipalaYPN25}. The latter: (\textit{i}) adopts one-shot prompting to enable step-by-step reasoning about program execution, and (\textit{ii}) provides a finite set of feasible options for code coverage (FOCC) for the model to select from, thus transforming prediction task into a constrained selection problem.

Figure~\ref{fig:fig-5} compares the performance of CRISPE, its ablation baselines (w/o one-shot prompting and w/o FOCC), and our prompt for execution reasoning. CRISPE yields consistently high performance across datasets for both GPT-5 Mini and Grok-4 Reasoning, as indicated by narrower performance variability. The ablation results further reveal that Grok-4 Reasoning relies more on FOCC than GPT-5 Mini: removing FOCC results in a substantially larger drop in $\mathrm{pass@}1$ for Grok-4 Reasoning (677.6\%) compared to GPT-5 Mini (25\%). This suggests that Grok-4 Reasoning is more sensitive to increased prompt complexity and subsequently longer step-by-step reasoning traces.



\vspace{-2.5pt}
\paragraph{Robustness to Counterfactual Path Selection Strategies.}
In Section~\ref{sec:datasets}, we selected the counterfactual path that yields the \textit{most} increase in code coverage relative to the original execution path. To evaluate the sensitivity of counterfactual reasoning in \tool to this selection heuristic, we compare against two additional strategies with the counterfactual path: (\textit{i}) picked at \textit{random}, and (\textit{ii}) results in the \textit{least} increase in code coverage.

\input{tables/sensitivity-backward}

Table~\ref{tab:table-3} reports the performance of GPT-5 Mini and Grok-4 Reasoning in counterfactual reasoning. We observe that the models are generally sensitive to the counterfactual objective, which determines the degree of causal reasoning required to identify input perturbations that induce a specific execution flow. In particular, paths that yield maximal coverage increase typically execute the core logic of the program, while those with minimal coverage increase often correspond to early exit conditions. The intermediate program state triggering the latter depends on satisfying highly specific constraints, thus requiring deeper causal understanding.
Between both models, we see that Grok-4 Reasoning is more robust to counterfactual path selection and records high performance across target branches, whereas GPT-5 Mini shows larger variability. As such, while a maximal coverage heuristic is reasonable, these findings suggest that augmenting \tool with additional counterfactual targets could enable an even comprehensive assessment of counterfactual reasoning performance.

\input{sections/qualitative}

%% file: diagrams/direction2.tex
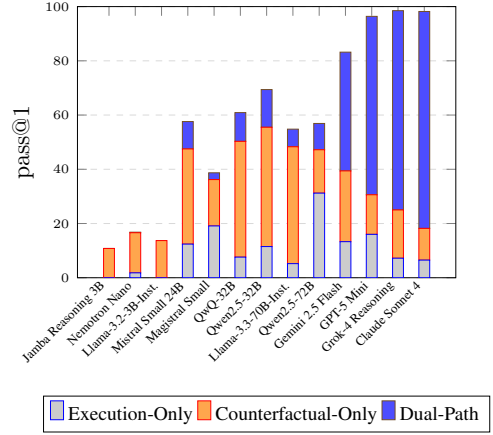
\begin{figure}[t]
\centering
\resizebox{0.85\linewidth}{!}{
\begin{tikzpicture}
\begin{axis}[
    ybar stacked,
    width=\linewidth,
    height=6cm,
    ymin=0, ymax=100,
    ylabel={$\mathrm{pass@}1$},
    ylabel style={yshift=-10pt, font=\normalsize},
    xtick=data,
    bar width=5pt,
    symbolic x coords={
        Jamba Reasoning 3B,
        Nemotron Nano,
        Llama-3.2-3B-Inst.,
        Mistral Small 24B,
        Magistral Small,
        QwQ-32B,
        Qwen2.5-32B,
        Llama-3.3-70B-Inst.,
        Qwen2.5-72B,
        Gemini 2.5 Flash,
        GPT-5 Mini,
        Grok-4 Reasoning,
        Claude Sonnet 4
    },
    xtick=data,
    x tick label style={
        rotate=45,
        anchor=east,
        font=\tiny
    },
    y tick label style={font=\tiny},
    ymajorgrids,
    grid style={dashed, gray!30},
    legend style={
        at={(0.5,-0.435)},
        anchor=north,
        legend cell align=center,
        legend columns = 3,
        fill=white,
        font=\small
    },
    every legend image/.style={draw}
]

\addplot+[fill=gray!40] coordinates {
    (Jamba Reasoning 3B, 0.0)
    (Nemotron Nano, 1.8)
    (Llama-3.2-3B-Inst., 0)
    (Mistral Small 24B, 12.4)
    (Magistral Small, 19.1)
    (QwQ-32B, 7.6)
    (Qwen2.5-32B, 11.5)
    
    (Llama-3.3-70B-Inst., 5.2)
    (Qwen2.5-72B, 31.2)

    (Gemini 2.5 Flash, 13.3)
    (GPT-5 Mini, 16.0)
    (Grok-4 Reasoning, 7.2)
    (Claude Sonnet 4, 6.5)
};

\addplot+[fill=orange!70] coordinates {
    (Jamba Reasoning 3B, 10.8)
    (Nemotron Nano, 14.8)
    (Llama-3.2-3B-Inst., 13.7)
    (Mistral Small 24B, 35.1)
    (Magistral Small, 17.1)
    (QwQ-32B, 42.7)
    (Qwen2.5-32B, 44.0)

    (Llama-3.3-70B-Inst., 43.1)
    (Qwen2.5-72B, 16.0)

    (Gemini 2.5 Flash, 26.1)
    (GPT-5 Mini, 14.6)
    (Grok-4 Reasoning, 17.8)
    (Claude Sonnet 4, 11.7)
};

\addplot+[fill=blue!70] coordinates {
    (Jamba Reasoning 3B, 0.0)
    (Nemotron Nano, 0.2)
    (Llama-3.2-3B-Inst., 0.0)
    (Mistral Small 24B, 10.1)
    (Magistral Small, 2.5)
    (QwQ-32B, 10.6)
    (Qwen2.5-32B, 13.9)

    (Llama-3.3-70B-Inst., 6.5)
    (Qwen2.5-72B, 9.7)
    
    (Gemini 2.5 Flash, 43.8)
    (GPT-5 Mini, 65.8)
    (Grok-4 Reasoning, 73.5)
    (Claude Sonnet 4, 80)
};

\legend{Execution-Only, Counterfactual-Only, Dual-Path}

\end{axis}
\end{tikzpicture}
}
\caption{Strict performance comparison ($\mathrm{pass@}1$, in \%) in execution-only, counterfactual-only, and dual-path reasoning evaluation settings across three datasets.}
\label{fig:quadrant-stacked}
\end{figure}

%% file: diagrams/relaxed.tex
\begin{figure}[t]
\centering
\resizebox{0.85\linewidth}{!}{
\begin{tikzpicture}
\begin{axis}[
    width=\linewidth,
    height=6cm,
    ymin=0, ymax=100,
    ylabel={$\mathrm{pass@}1$},
    ylabel style={yshift=-10pt, font=\normalsize},
    xtick=data,
    symbolic x coords={
        Jamba Reasoning 3B,
        Nemotron Nano,
        Llama-3.3-3B-Inst.,
        Mistral Small 24B,
        Magistral Small, QwQ-32B,
        Qwen2.5-32B,
        Llama-3.3-70B-Inst., 
        Qwen2.5-72B, 
        Gemini 2.5 Flash, 
        GPT-5 Mini,
        Grok-4 Reasoning,
        Claude Sonnet 4
    },
    x tick label style={
        rotate=45,
        anchor=east,
        font=\tiny
    },
    y tick label style={font=\tiny},
    ymajorgrids,
    grid style={dashed, gray!30},
    legend style={
        at={(0.5,-0.435)},
        anchor=north,
        legend columns=2,
        fill=white,
        font=\small
    },
    every legend image/.style={draw}
]

\addplot[name path=fwd-low, draw=none, forget plot] coordinates {
    (Jamba Reasoning 3B, 0.0)
    (Nemotron Nano, 0.0)
    (Llama-3.3-3B-Inst., 0.0)
    (Mistral Small 24B, 2.9)
    (Magistral Small, 3.3)
    (QwQ-32B, 1.8)
    (Qwen2.5-32B, 0.0)
    (Llama-3.3-70B-Inst., 0.0)
    (Qwen2.5-72B, 2.9)
    (Gemini 2.5 Flash, 9.8)
    (GPT-5 Mini, 40.0) 
    (Grok-4 Reasoning, 57.2)
    (Claude Sonnet 4, 61.5)
};

\addplot[name path=fwd-high, draw=none, forget plot] coordinates {
    (Jamba Reasoning 3B, 0.0)
    (Nemotron Nano, 2.7)
    (Llama-3.3-3B-Inst., 0.0)
    (Mistral Small 24B, 27.5)
    (Magistral Small, 26.4)
    (QwQ-32B, 25.6)
    (Qwen2.5-32B, 32.5)
    (Llama-3.3-70B-Inst., 15.0)
    (Qwen2.5-72B, 27.5)
    (Gemini 2.5 Flash, 72.7)
    (GPT-5 Mini, 96.8) 
    (Grok-4 Reasoning, 88.7)
    (Claude Sonnet 4, 94.9)
};

\addplot[
    fill=magenta!50,
    fill opacity=0.225,
    draw=none,
    forget plot
] fill between[of=fwd-low and fwd-high];

\addplot[
    color=magenta,
    mark=o,
    mark size=2pt,
    thick,
    forget plot
] coordinates {
    (Jamba Reasoning 3B, 0.0)
    (Nemotron Nano, 1.0)
    (Llama-3.3-3B-Inst., 0.0)
    (Mistral Small 24B, 15.2)
    (Magistral Small, 14.8)
    (QwQ-32B, 13.7)
    (Qwen2.5-32B, 15.1)
    (Llama-3.3-70B-Inst., 6.9)
    (Qwen2.5-72B, 15.2)
    (Gemini 2.5 Flash, 41.3)
    (GPT-5 Mini, 68.4) 
    (Grok-4 Reasoning, 73.0)
    (Claude Sonnet 4, 78.2)
};

\addplot[name path=ovr-low, draw=none, forget plot] coordinates {
    (Jamba Reasoning 3B, 1.7)
    (Nemotron Nano, 29.0)
    (Llama-3.3-3B-Inst., 40.1)
    (Mistral Small 24B, 62.4)
    (Magistral Small, 59.7)
    (QwQ-32B, 12.2)
    (Qwen2.5-32B, 66.7)
    (Llama-3.3-70B-Inst., 68.7)
    (Qwen2.5-72B, 66.9)
    (Gemini 2.5 Flash, 34.7)
    (GPT-5 Mini, 44.6) 
    (Grok-4 Reasoning, 97.1)
    (Claude Sonnet 4, 95.2) 
};

\addplot[name path=ovr-high, draw=none, forget plot] coordinates {
    (Jamba Reasoning 3B, 7.6)
    (Nemotron Nano, 40.4)
    (Llama-3.3-3B-Inst., 44.0)
    (Mistral Small 24B, 64.6)
    (Magistral Small, 67.8)
    (QwQ-32B, 55.3)
    (Qwen2.5-32B, 80.5)
    (Llama-3.3-70B-Inst., 82.7)
    (Qwen2.5-72B, 84.3)
    (Gemini 2.5 Flash, 98.6)
    (GPT-5 Mini, 100.0) 
    (Grok-4 Reasoning, 98.1)
    (Claude Sonnet 4, 98.8)
};

\addplot[
    fill=teal!50,
    fill opacity=0.275,
    draw=none,
    forget plot
] fill between[of=ovr-low and ovr-high];

\addplot[
    color=teal!70,
    mark=*,
    mark size=2pt,
    thick,
    forget plot
] coordinates {
    (Jamba Reasoning 3B, 4.6)
    (Nemotron Nano, 34.7)
    (Llama-3.3-3B-Inst., 42.0)
    (Mistral Small 24B, 63.5)
    (Magistral Small, 63.8)
    (QwQ-32B, 33.8)
    (Qwen2.5-32B, 73.6)
    (Llama-3.3-70B-Inst., 75.7)
    (Qwen2.5-72B, 75.6)
    (Gemini 2.5 Flash, 66.7)
    (GPT-5 Mini, 75.9)
    (Grok-4 Reasoning, 97.6)
    (Claude Sonnet 4, 97.0)
};
\addlegendimage{line legend, color=magenta, mark=o}
\addlegendentry{Strict (Exact Match)}
\addlegendimage{line legend, color=teal, mark=*}
\addlegendentry{Relaxed (Jaccard Similarity)}

\end{axis}
\end{tikzpicture}
}
\caption{Strict vs.\ relaxed performance comparison ($\mathrm{pass@}1$, in \%) on execution reasoning $\mathcal{R}_{exec}$. Here, shaded regions indicate $\pm$std across three datasets.}
\label{fig:strict-relaxed}
\end{figure}
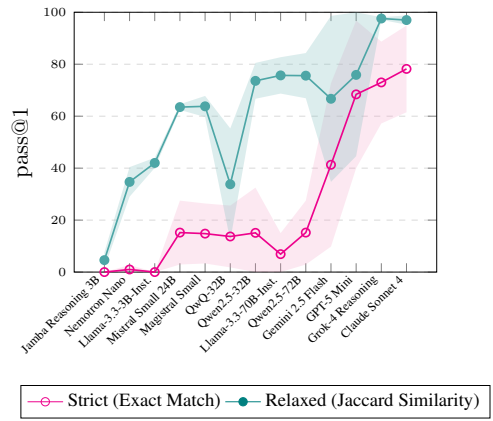

%% file: diagrams/ablation.tex
\begin{table}[t]
\centering
\setlength{\tabcolsep}{4pt} 
\resizebox{\linewidth}{!}{
\begin{tabular}{c|c|cc|cc|cc}
\toprule
\multicolumn{1}{c|}{\multirow{2}{*}{\makecell{\textbf{Dataset ($\rightarrow$)} \\ \textbf{Model ($\downarrow$)}}}} & \multicolumn{1}{c|}{\multirow{2}{*}{\makecell{\textbf{C/F} \\ \textbf{Reasoning}}}}                               & \multicolumn{2}{c|}{\textbf{CruxEval}}                                                       & \multicolumn{2}{c|}{\textbf{HumanEval}}                                                      & \multicolumn{2}{c}{\textbf{PythonSaga}}                                        \\ 
\cmidrule{3-8} 
\multicolumn{1}{c|}{}                                & & $k\!=\!1$ & $k\!=\!5$ & $k\!=\!1$ & $k\!=\!5$ & $k\!=\!1$ & $k\!=\!5$ \\
\midrule
\multirow{2}{*}{GPT-5 Mini} & Generation & 89.3 & {\cellcolor{gray!12}{97.7}} & 93.0 & {\cellcolor{gray!12}{99.0}} & 87.2 & {\cellcolor{gray!12}{95.7}} \\
                                                 & Mutation & 81.2 & {\cellcolor{gray!12}{97.0}} & 77.0 & {\cellcolor{gray!12}{88.0}} & 83.0 & {\cellcolor{gray!12}{95.7}} \\
\hline
\multirow{2}{*}{\begin{tabular}{c} Grok-4 \\ Reasoning \end{tabular}} & Generation & 97.3 & {\cellcolor{gray!12}{99.0}} & 98.0 & {\cellcolor{gray!12}{99.0}} & 93.6 & {\cellcolor{gray!12}{95.7}} \\
                                                 & Mutation & 93.0 & {\cellcolor{gray!12}{97.0}} & 85.0 & {\cellcolor{gray!12}{88.0}} & 93.6 & {\cellcolor{gray!12}{95.7}} \\
\bottomrule
\end{tabular}
}
\caption{\textbf{Ablation Study.} Model performance under input generation vs. mutation (\textit{ours}) for counterfactual reasoning ($\mathrm{pass@}k$, in $\%$) across three datasets.}
\label{tab:ablation}
\end{table}

%% file: diagrams/sensitivity-forward.tex
\begin{figure}[t]
\centering
\resizebox{0.85\linewidth}{!}{
\begin{tikzpicture}
\begin{axis}[
    width=\linewidth,
    height=6cm,
    ybar,
    bar width=8pt,
    ymin=0, ymax=100,
    ylabel={$\mathrm{pass@}1$},
    ylabel style={yshift=-10pt, font=\normalsize},
    symbolic x coords={
        CRISPE,
        {w/o 1-shot},
        {w/o FOCC},
        Simple (\textit{ours})
    },
    xtick=data,
    x tick label style={font=\tiny},
    y tick label style={font=\tiny},
    legend style={
        at={(0.5,-0.15)},
        anchor=north,
        legend columns=2,
        column sep=0.15cm,
        draw=black,
        fill=white,
        font=\small
    },
    ymajorgrids,
    grid style={dashed, gray!30},
    enlarge x limits=0.2
]

\addplot+[
    pattern=north east lines,
    pattern color = blue!30,
    error bars/.cd,
        y dir=both,
        y explicit,
] coordinates {
    (CRISPE, 85.1) +- (0,8) 
    ({w/o 1-shot}, 88.3) +- (0,6.66)
    ({w/o FOCC}, 68.1) +- (0,14.42)
    (Simple (\textit{ours}), 68.4) +- (0,28.3) 
};

\addplot+[
    pattern=dots,
    pattern color=red!40,
    error bars/.cd,
        y dir=both,
        y explicit,
] coordinates {
    (CRISPE, 97.2) +- (0,4.9) 
    ({w/o 1-shot}, 91.6) +- (0,4.4)
    ({w/o FOCC}, 12.5) +- (0, 12.5)
    (Simple (\textit{ours}), 72.3) +- (0,16.89) 
};

\legend{GPT-5 Mini, Grok-4 Reasoning}

\end{axis}
\end{tikzpicture}
}
\caption{
\textbf{Robustness to prompt complexity.}
Average model performance for execution reasoning ($\mathrm{pass@}1$, in \%). Here, whiskers denote $\pm\sigma$ across three datasets.
}
\label{fig:fig-5}
\end{figure}
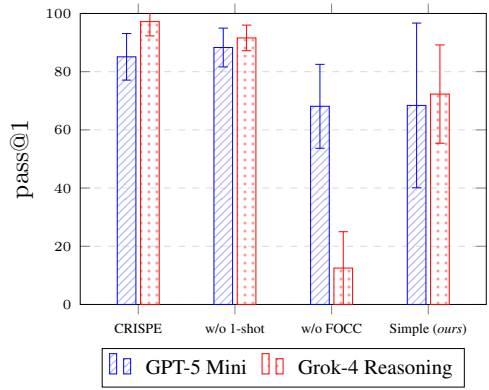

%% file: tables/sensitivity-backward.tex
\begin{table}[t]
\centering
\setlength{\tabcolsep}{4pt} 
\resizebox{\linewidth}{!}{
\begin{tabular}{c|c|cc|cc|cc}
\toprule
\multicolumn{1}{c|}{\multirow{2}{*}{\makecell{\textbf{Dataset ($\rightarrow$)} \\ \textbf{Model ($\downarrow$)}}}} & \multicolumn{1}{c|}{\multirow{2}{*}{\makecell{\textbf{C/F Path} \\ \textbf{Selection}}}}                               & \multicolumn{2}{c|}{\textbf{CruxEval}}                                                       & \multicolumn{2}{c|}{\textbf{HumanEval}}                                                      & \multicolumn{2}{c}{\textbf{PythonSaga}}                                        \\ 
\cmidrule{3-8} 
\multicolumn{1}{c|}{}                                & & $k\!=\!1$ & $k\!=\!5$ & $k\!=\!1$ & $k\!=\!5$ & $k\!=\!1$ & $k\!=\!5$ \\
\midrule
\multirow{3}{*}{GPT-5 Mini} & Random & 71.4 & {\cellcolor{gray!12}{83.3}} & 67.9 & {\cellcolor{gray!12}{67.9}} & 58.8 & {\cellcolor{gray!12}{70.6}} \\
                                                 & Min & 76.2 & {\cellcolor{gray!12}{92.9}} & 71.4 & {\cellcolor{gray!12}{89.3}} & 76.5 & {\cellcolor{gray!12}{94.1}} \\
                                                 & Max (\textit{ours}) & 85.7 & {\cellcolor{gray!12}{97.6}} & 78.6 & {\cellcolor{gray!12}{89.3}} & 60.7 & {\cellcolor{gray!12}{94.1}} \\
\hline
\multirow{3}{*}{\begin{tabular}{c} Grok-4 \\ Reasoning \end{tabular}} & Random & 85.7 & {\cellcolor{gray!12}{88.1}} & 89.3 & {\cellcolor{gray!12}{89.3}} & 94.1 & {\cellcolor{gray!12}{94.1}} \\
                                                 & Min & 85.7 & {\cellcolor{gray!12}{90.5}} & 89.3 & {\cellcolor{gray!12}{89.3}} & 94.1 & {\cellcolor{gray!12}{94.1}} \\
                                                 & Max (\textit{ours}) & 97.6 & {\cellcolor{gray!12}{97.6}} & 89.3 & {\cellcolor{gray!12}{89.3}} & 88.2 & {\cellcolor{gray!12}{94.1}} \\
\bottomrule
\end{tabular}
}
\caption{\textbf{Robustness to counterfactual paths.} Model performance on counterfactual reasoning ($\mathrm{pass@}k$, $\%$).}
\label{tab:table-3}
\end{table}

%% file: sections/qualitative.tex
\subsection{Qualitative Analysis of Model Outputs}

To identify the root causes of model failures, we manually inspected reasoning traces produced by closed-source models under both execution (forward) and counterfactual (backward) reasoning. Following prior work~\cite{DBLP:journals/corr/abs-2512-00215}, we categorize failures into following high-level error types: \textit{Computation Errors}, \textit{Control Flow Errors}, \textit{Lack of Fact Verification}, \textit{Native API Misevaluation}, \textit{Prompt Misread}, \textit{Null Response}, \textit{Output Format Errors}, \textit{Misreporting Final Output}, \textit{Skipping Statements}, and \textit{Hallucinations} (Appendix~\ref{app:qualitative}).

As shown in Table~\ref{tab:error-tax}, the error categories do not apply uniformly across both reasoning tasks. For execution reasoning, the model errors primarily occur due to: misunderstanding task instructions (Prompt Misread), incorrectly parsing and processing program representations (Skipping Statements, Hallucinations), and failing to produce outputs (Null Response). For counterfactual reasoning, Gemini-2.5 Flash and GPT-5 Mini make the most errors due to producing a null response or an error in the output format. In contrast, Grok-4 Reasoning and Claude Sonnet 4 produce few or no such errors. However, Claude Sonnet 4 exhibits more semantic errors (\textit{e.g.}, Native API Misevaluation, Control Flow Errors) than even Gemini-2.5 Flash, suggesting a gap in modeling execution flow which requires the model to maintain causal consistency. 

\vspace{-2.5pt}
\paragraph{Illustration.}
In multiple API-centric tasks that involve string manipulation, Claude Sonnet 4 correctly predicts the code coverage (\textit{i.e.}, execution reasoning) but fails to produce a valid counterfactual input (\textit{i.e.}, counterfactual reasoning).
In some cases, 
the failure arises from a misinterpretation of API semantics (\textit{e.g.}, {\small \texttt{splitlines()}} vs.\ {\small\texttt{split()}}). Consistent with recent findings on Claude's handling of {\small\texttt{split()}} API~\cite{DBLP:journals/corr/abs-2512-00215}, these are not due to ambiguity in task specification but reflect an incorrect understanding of API behavior.


The above analyses show that models may succeed in execution reasoning while failing in counterfactual reasoning due to semantic errors, or vice versa. As a result, isolated evaluations based on execution-only or counterfactual-only settings do not provide a robust evaluation of model capabilities, underscoring the need for jointly evaluating both directions, as in our dual-path framework.

%% file: sections/related-short.tex
Emergent reasoning capabilities in LLMs have been studied across multiple domains, including mathematical reasoning~\cite{DBLP:journals/corr/abs-2401-05384,DBLP:conf/acl/JainMTS25,DBLP:conf/nips/LewkowyczADDMRS22}, formal logic~\cite{DBLP:conf/icml/MorishitaMYS23}, and program understanding~\cite{DBLP:journals/corr/abs-2507-05269,DBLP:conf/icse/ChenP000025,DBLP:journals/corr/abs-2512-00215}. For source code, prior work has focused on evaluating the ability of LLMs to reason about program properties, either derived from program structure and semantics or observed runtime behaviors, \textit{i.e.}, static~\cite{DBLP:journals/pacmpl/YadavallyLWN24,DBLP:journals/corr/abs-2507-05269} and dynamic code reasoning~\cite{DBLP:conf/acl/LiuLCJSFSD23,DBLP:journals/pacmse/YadavallyLN24,DBLP:conf/icse/PatelYDN25}, respectively. 
In this work, we focus on fine-grained benchmarking of dynamic code understanding.


CruxEval~\cite{DBLP:conf/icml/GuRLSS024} evaluates LLMs by pairing output prediction (given a program and input) with input prediction (given a program and output). CodeMind~\cite{DBLP:journals/corr/abs-2402-09664} extends beyond input-output prediction by introducing tasks on understanding semantics, specification, and execution simulation. R-Eval~\cite{DBLP:conf/icse/ChenP000025} introduced IC-Score, a measure derived from the composition of performance on program execution-specific tasks such as predicting code coverage, program states, execution paths, and program outputs. 
However, these benchmarks remain limited to \textit{single-path evaluations} induced by specific inputs.
While CES~\cite{DBLP:journals/corr/abs-2510-15079} focuses on reasoning consistency across multiple inputs, it treats resulting execution paths independently, 
thereby leaving the causal logic that is characteristic of program execution unexplored.
In contrast, \tool introduces a dual-path reasoning framework. By requiring the model to predict the observed behaviors along an execution path, and the behavioral changes required to induce an alternative flow, \tool provides a more robust and systematic evaluation of LLMs over program execution.


%% file: sections/conclusion.tex

In this work, we argue that reasoning about program execution is inherently dual in nature, requiring both the ability to predict dynamic program behavior and to reason counterfactually to induce alternative behaviors. Our results show that strong performance on either task does not consistently translate to success under joint evaluation, an asymmetry not reflected in single-path evaluations. These findings suggest that the proposed dual-path reasoning framework provides a more robust evaluation of dynamic code understanding in LLMs and can be extended to alternative pairs of observed runtime behaviors and counterfactual objectives.

%% file: sections/limitation.tex
\paragraph{Data Contamination.}
This work sources Python programs from public benchmarks, \textit{i.e.}, CruxEval, HumanEval, and PythonSaga, and therefore may be susceptible to data contamination. However, the formulation in \tool mitigates this risk by requiring fine-grained reasoning over intermediate program state rather than direct input-output mapping. In particular, the counterfactual reasoning task requires conditional mutation of inputs, where a model must identify and analyze specific branches and perturb inputs to redirect the flow of execution. Thus, even if a model has potentially ``seen'' the underlying program, successfully synthesizing a constraint-satisfying input requires an explicit understanding of how input perturbations propagate under execution. As a result, model performance in \tool is less likely to stem from surface-level memorization and instead reflects its ability to reason over intrinsic execution flow.


\paragraph{Programming Languages.} Constructing our benchmark relies on the availability of tools for execution tracing and instrumentation to reliably capture fine-grained runtime information. While these are readily available for Python (\textit{e.g.}, SlipCover), extending this to other programming languages requires substantial engineering effort to build secure and reproducible sandboxed runtime environments. As a result, \tool focuses on Python programs drawn from widely adopted CruxEval, HumanEval, and PythonSaga datasets. However, the proposed dual-path reasoning framework itself is programming language-agnostic.

\paragraph{Counterfactual Path Selection.}
By default, this work selects the branch that maximizes increase in code coverage relative to the original execution. 
%
%
However, this can bias selection towards counterfactual paths that diverge more from the original execution, while overlooking paths corresponding to more localized behavioral variations.
Although we explore alternative strategies in Section~\ref{sec:sensitivity}, identifying counterfactual paths that are both informative and representative remains challenging.  Therefore, model performance reported in \tool may vary under different counterfactual selection strategies, and our findings should be interpreted in the context of this design choice.

\paragraph{Dynamic Program Behavior and Counterfactual Objective.} This work focuses on evaluating dynamic program behavior through code coverage prediction, with the corresponding counterfactual objective being branch-targeted input mutation. While this provides a concrete and tractable formulation of dual-path reasoning, other alternatives are possible. For example, execution reasoning could instead target predicting program outputs or intermediate execution states, while counterfactual reasoning might involve identifying inputs that induce specific outputs or state transitions.~Exploring such alternatives are useful to coding tasks like testing, debugging, and may yield complementary evaluation perspectives. This is left for future work.
%
%

%% file: sections/appendix-final.tex
\section{Prompt Design}
\label{sec:prompt-design}

We design dedicated prompts to evaluate distinct reasoning abilities of LLMs:
\begin{itemize}[itemsep=0pt]
    \item Figure~\ref{fig:prompt-combined} (\textit{top}): Forward reasoning, where the model must simulate execution of a program to predict which lines would be executed for a given input (\textit{i.e.}, statement coverage).

    \item Figure~\ref{fig:prompt-combined} (\textit{bottom}): Backward reasoning, where the model must predict an input that would reach a branch uncovered by the original input for the same program.
\end{itemize}
Both templates enforce strict formatting to enable automated parsing and evaluation. Here, \verb|[PYTHON]|/\verb|[/PYTHON]| tags enclose the line number-indexed Python program, \verb|[TASK]|/\verb|[/TASK]| tags enclose the corresponding task instructions, \verb|<REASONING>|/\verb|</REASONING>| tags enclose model-generated reasoning traces, and \verb|[ANSWER]|/\verb|[/ANSWER]| tags enclose the model-generated task outputs.

\input{diagrams/prompt-reformatted}

\section{Counterfactual Path Selection}
\label{app:line-priority}

In this section, we describe the algorithm used to define behavioral goals for branch-targeted input mutation under counterfactual reasoning ($\mathcal{R}_{cf}$). As noted in Section~\ref{sec:datasets}, this procedure grounds counterfactual input mutation in branches that yield maximal increase in code coverage.

Given a coverage map produced by executing the program on a seed input, we first identify the set of uncovered source lines. We then analyze the program’s abstract syntax tree (AST) to enumerate all conditional branches and loop constructs, associating each such construct with the set of source lines contained in its body. The primary objective is to identify branches whose bodies are \textit{entirely uncovered}, all of which represent possible targets for counterfactual exploration. Among these, we select the branch with the largest body, measured as the number of statements it contains. This favors branches that are likely to yield the \textit{highest} increase in code coverage if successfully executed.

If no fully uncovered branch exists, we select uncovered lines that do not belong to the body of any conditional or loop. These standalone lines typically correspond to return statements, computations, or other control-flow–independent code regions. In this case, we deterministically select the uncovered line with the smallest line number to serve as the target.
%
%
Finally, if no such branches or statements can be identified, we ignore such programs as no meaningful counterfactual target is available under the current state. 
We formalize this alternative path selection process as follows:

\input{diagrams/algorithm}

\section{Qualitative Analysis of Model Outputs}

\subsection{Error Taxonomy}\label{app:qualitative}

We manually inspected reasoning traces for failure cases and categorized them into ten distinct error types, as summarized in Table~\ref{tab:error-tax}.

\input{tables/qual-categories-expanded}

\subsection{Illustrative Examples}\label{app:examples}

\begin{itemize}[itemsep=0pt]
    \item In Figure~\ref{fig:example-1} (CRUXEval-492), the model produces a largely correct trace but fails due to a basic counting error (hallucinating character counts), demonstrating that even strong models are prone to verifiable logic lapses when constraints are inverted.
    \item In Figure~\ref{fig:example-2}, the model misinterprets the behavior of Python's \texttt{splitlines()}, propagating this error into an invalid input suggestion. This aligns with discrepancies noted in recent literature regarding Claude's handling of the \texttt{split()} function~\cite{DBLP:journals/corr/abs-2512-00215}. 
\end{itemize}

\input{diagrams/Example1}

\input{diagrams/Example2}

%% file: diagrams/prompt-reformatted.tex
\begin{figure*}
    \centering
\begin{tcolorbox}[colback=gray!10, colframe=black, title=\textbf{Prompt Template for Execution (Forward) Reasoning}]
You are given a Python program inside the [PYTHON] and [/PYTHON] tags and a single task between [TASK][/TASK] tags. Your only function is to comply with Output requirements below. \\[0.5em]

[PYTHON] \\[0.3em]
\textbf{\{program\_code\}} \\[0.3em]
[/PYTHON] \\[0.5em]

[TASK] \\[0.3em]
Analyze the provided Python script to determine its code coverage when executed. Identify all executable line numbers that are run at least once. \\[0.3em]
[/TASK] \\[0.5em]

Output requirements (strict):

1. First, analyze the execution paths and predict which lines would be executed. Put your reasoning inside <REASONING></REASONING> tags.

2. Then, provide the final answer inside [ANSWER] and [/ANSWER] tags as a single JSON object with one key:

   - \texttt{executed\_lines}: JSON list of integers (sorted ascending) for all executable lines that would run.
\end{tcolorbox}

\vspace{6pt}

\begin{tcolorbox}[colback=gray!10, colframe=black, title=\textbf{Prompt Template for Counterfactual (Backward) Reasoning}]
You are an expert in directed fuzzing and test-case generation. You are given a Python program inside the [PYTHON] and [/PYTHON] tags, and a single task between [TASK][/TASK] tags. Your only function is to comply with Output requirements below. \\[0.5em]

[PYTHON] \\[0.3em]
\textbf{\{program\_code\}} \\[0.3em]
[/PYTHON] \\[0.5em]

[TASK] \\[0.3em]
The target line \textbf{\{priority\_line\}} is either a conditional (if/elif), a loop header (for/while), or a return statement. Mutate the existing input seed so that, when passed to the function under test, the body of the conditional or loop at line \textbf{\{priority\_line\}} will be executed. If the line is a return statement, that line should be triggered. \\[0.3em]
[/TASK] \\[0.5em]

Output requirements (strict):

1. First, analyze the program and reason about what inputs would reach line \textbf{\{priority\_line\}}. Put your reasoning inside <REASONING></REASONING> tags.

2. Then, provide the final answer inside [ANSWER] and [/ANSWER] tags with **only** the comma-separated argument values (for example: \verb|`["a","bb"], 0.5`|).
\end{tcolorbox}

    \caption{Prompt templates for: (\textit{top}) execution, and (\textit{bottom}) counterfactual reasoning in \tool.}
    \label{fig:prompt-combined}
\end{figure*}

%% file: diagrams/algorithm.tex
{
\begin{algorithm}
\caption{Alternative Path Selection}
\label{alg:target-selection}
\DontPrintSemicolon
\SetAlgoLined
\KwIn{Coverage map $\mathcal{C}$, AST $\mathcal{A}$}
\KwOut{Target line number $\ell^\ast$ or \texttt{null}}
$\mathcal{U} \leftarrow \{\ell \mid \text{coverage\_count}(\ell) = 0 \}$\;
$\mathcal{T} \leftarrow \emptyset$\;
\ForEach{branch $b \in$ \textsc{FindAllBranches}($\mathcal{A}$)}{
    $\mathcal{B} \leftarrow$ \textsc{LinesInBranchBody}($b$)\;
    \If{$\mathcal{B} \neq \emptyset$ \textbf{and} $\mathcal{B} \subseteq \mathcal{U}$}{
        $\mathcal{T} \leftarrow \mathcal{T} \cup \{b\}$\;

    }
}
\If{$\mathcal{T} \neq \emptyset$}{
    $\mathcal{S} \leftarrow \emptyset$\;
    \ForEach{branch $b \in \mathcal{T}$}{
        $s_b \leftarrow |\textsc{LinesInBranchBody}(b)|$\;
        $\mathcal{S} \leftarrow \mathcal{S} \cup \{(b, s_b)\}$\;
    }
    $b^\ast \leftarrow \arg\max_{(b, s_b) \in \mathcal{S}} s_b$\;
    \Return $\textsc{LineNumber}(b^\ast)$\;
}
$\mathcal{U}$\textsubscript{standalone} $\leftarrow \mathcal{U} \setminus \textsc{AllBranchBodyLines}(\mathcal{A})$\;
\If{$\mathcal{U}_{\text{standalone}} \neq \emptyset$}{
    \Return $\min(\mathcal{U}_{\text{standalone}})$\;
}
\Return \texttt{null}\;
\end{algorithm}
}

%% file: tables/qual-categories-expanded.tex
\begin{table*}
\centering
\setlength{\tabcolsep}{4pt} 
\resizebox{\linewidth}{!}{
\begin{tabular}{p{11.75cm}|c|cccc}
\toprule
\makecell{\textbf{Error Category}}   & \multirow{1}{*}{\textbf{Reasoning}} & 
\makecell{Gemini-\\2.5 Flash} & \makecell{GPT-5\\Mini} & \makecell{Grok-4\\Reasoning} & \makecell{Claude\\Sonnet-4} \\ 
\midrule
\multirow{2}{*}{\parbox{11.75cm}{\textit{Computation Errors}: Arithmetic and type conversion faults}}        & $\mathcal{R}_{exec}$                           & 3                                                                   & 2                                                             & 0                                                                   & 2                                                                    \\
                                           & {\cellcolor{gray!12}{$\mathcal{R}_{cf}$}}                      & {\cellcolor{gray!12}{2}}                                                                   & {\cellcolor{gray!12}{0}}                                                             & {\cellcolor{gray!12}{2}}                                                                   & {\cellcolor{gray!12}{2}}                                                                    \\ \midrule
\multirow{2}{*}{\parbox{11.75cm}{\textit{Control Flow Errors}: Incorrect predicate evaluation in loops or conditionals}}       & $\mathcal{R}_{exec}$                           & 3                                                                   & 0                                                             & 0                                                                   & 1                                                                    \\
                                           & {\cellcolor{gray!12}{$\mathcal{R}_{cf}$}}                      & {\cellcolor{gray!12}{1}}                                                                   & {\cellcolor{gray!12}{1}}                                                             & {\cellcolor{gray!12}{2}}                                                                   & {\cellcolor{gray!12}{4}}                                                                    \\ \midrule
\multirow{2}{*}{\parbox{11.75cm}{\textit{Lack of Fact Verification}: Model fails to logically link input mutation to target path, leading to a false result despite seemingly coherent reasoning steps}} & $\mathcal{R}_{exec}$                           & 0                                                                   & 1                                                             & 1                                                                   & 5                                                                    \\
                                           & {\cellcolor{gray!12}{$\mathcal{R}_{cf}$}}                      & {\cellcolor{gray!12}{1}}                                                                   & {\cellcolor{gray!12}{3}}                                                             & {\cellcolor{gray!12}{3}}                                                                   & {\cellcolor{gray!12}{3}}                                                                    \\ \midrule
\multirow{2}{*}{\parbox{11.75cm}{\textit{Native API Misevaluation}: Hallucinations or errors related to standard library behavior (\textit{e.g.}, string manipulation)}}  & $\mathcal{R}_{exec}$                           & 1                                                                   & 0                                                             & 0                                                                   & 1                                                                    \\
                                           & {\cellcolor{gray!12}{$\mathcal{R}_{cf}$}}                      & {\cellcolor{gray!12}{2}}                                                                   & {\cellcolor{gray!12}{1}}                                                             & {\cellcolor{gray!12}{1}}                                                                   & {\cellcolor{gray!12}{4}}                                                                    \\ \midrule
\multirow{2}{*}{\parbox{11.75cm}{\textit{Prompt Misread}: Model misinterprets the task objective, such as targeting the execution line number rather than the branch body}}            & $\mathcal{R}_{exec}$                           & 57                                                                  & 4                                                             & 7                                                                   & 4                                                                    \\
                                           & {\cellcolor{gray!12}{$\mathcal{R}_{cf}$}}                      & {\cellcolor{gray!12}{18}}                                                                  & {\cellcolor{gray!12}{0}}                                                             & {\cellcolor{gray!12}{5}}                                                                   & {\cellcolor{gray!12}{3}}                                                                    \\ \midrule
\multirow{2}{*}{\parbox{11.75cm}{\textit{Null Response}: Model generates empty or incomplete outputs}}             & $\mathcal{R}_{exec}$                           & 90                                                                  & 40                                                            & 0                                                                   & 2                                                                    \\
                                           & {\cellcolor{gray!12}{$\mathcal{R}_{cf}$}}                      & {\cellcolor{gray!12}{81}}                                                                  & {\cellcolor{gray!12}{33}}                                                            & {\cellcolor{gray!12}{0}}                                                                   & {\cellcolor{gray!12}{6}}                                                                    \\ \midrule
\multirow{2}{*}{\parbox{11.75cm}{\textit{Output Format Errors}: Test inputs under counterfactual reasoning  violate the function signature (\textit{e.g.}, providing a list instead of comma-separated values)}}      & $\mathcal{R}_{exec}$                           & 0                                                                   & 0                                                             & 0                                                                   & 0                                                                    \\
                                           & {\cellcolor{gray!12}{$\mathcal{R}_{cf}$}}                      & {\cellcolor{gray!12}{22}}                                                                  & {\cellcolor{gray!12}{39}}                                                            & {\cellcolor{gray!12}{11}}                                                                  & {\cellcolor{gray!12}{1}}                                                                    \\ \midrule
\multirow{2}{*}{\parbox{11.75cm}{\textit{Misreporting Final Output}: Model correctly infers which lines are executed but inconsistently reports them in its final output (or vice versa)}} & $\mathcal{R}_{exec}$                           & 1                                                                   & 1                                                             & 0                                                                   & 3                                                                    \\
                                           & {\cellcolor{gray!12}{$\mathcal{R}_{cf}$}}                      & {\cellcolor{gray!12}{0}}                                                                   & {\cellcolor{gray!12}{0}}                                                             & {\cellcolor{gray!12}{0}}                                                                   & {\cellcolor{gray!12}{0}}                                                                    \\ \midrule
\multirow{2}{*}{\parbox{11.75cm}{\textit{Skipping Statements}: Model omits one or more consecutively executed, executable statements within the same basic block}}       & $\mathcal{R}_{exec}$                           & 17                                                                  & 0                                                             & 1                                                                   & 19                                                                   \\
                                           & {\cellcolor{gray!12}{$\mathcal{R}_{cf}$}}                      & {\cellcolor{gray!12}{0}}                                                                   & {\cellcolor{gray!12}{0}}                                                             & {\cellcolor{gray!12}{0}}                                                                   & {\cellcolor{gray!12}{0}}                                                                    \\ \midrule
\multirow{2}{*}{\parbox{11.75cm}{\textit{Hallucination}: Model marks non-executable lines as executed}}             & $\mathcal{R}_{exec}$                           & 51                                                                  & 41                                                            & 79                                                                  & 46                                                                   \\
                                           & {\cellcolor{gray!12}{$\mathcal{R}_{cf}$}}                      & {\cellcolor{gray!12}{0}}                                                                   & {\cellcolor{gray!12}{0}}                                                             & {\cellcolor{gray!12}{0}}                                                                   & {\cellcolor{gray!12}{0}}                                                                    \\ \bottomrule
\end{tabular}
}
\caption{Error taxonomy for Gemini-2.5 Flash, GPT-5 Mini, Grok-4 Reasoning and Claude Sonnet-4 models across three datasets, under both execution ($\mathcal{R}_{exec}$) and counterfactual ( $\mathcal{R}_{cf}$) reasoning settings.}
\label{tab:error-tax}
\end{table*}

%% file: diagrams/Example1.tex
\begin{figure*}
    \centering
\begin{tcolorbox}[colback=white, colframe=black, boxrule=0.8pt]
\begin{minipage}{0.6\textwidth}
\lstset{style=exstyle}
\begin{lstlisting}[language=Python,numbers=left,numberstyle=\tiny,stepnumber=1,numbersep=8pt]
import unittest

def f(text, value):
    ls = list(text)
    if (ls.count(value)) % 2 == 0:
        while value in ls:
            ls.remove(value)
    else:
        ls.clear()
    return ''.join(ls)

# Test Assertion
unittest.TestCase().assertEqual(
    f('abbkebaniuuwrzvr', 'm'), 'abbkebaniuuwrzvr'
)
\end{lstlisting}
\end{minipage}
\hfill
\begin{minipage}{0.35\textwidth}
\textbf{Program:} CRUXEval-492 \\[4pt]
\textbf{Uncovered Branch:} 6$\rightarrow$7 \\[4pt]
\textbf{Target:} Branch at line 6 \\[6pt]

\textbf{LLM Answer:} \\
\texttt{`abbkebaniuuwrzvr', `b'} \\[6pt]

\textbf{Error:} Computation Error and Prompt Misunderstanding
\end{minipage}

\end{tcolorbox}

\vspace{6pt}

\begin{tcolorbox}[colback=gray!10, colframe=black, title=\textbf{Chain-of-Thought Reasoning Trace of Claude Sonnet 4}]
Let me analyze the program to understand when line 6 will be executed. The function \texttt{f(text, value)} takes two parameters: \texttt{text} is converted to a list \texttt{ls}, and \texttt{value} is the character to count and potentially remove.
The flow is:
\begin{enumerate}[topsep=0pt, itemsep=0pt]
    \item Convert text to list.
    \item Check if count of \texttt{value} is even.
    \item If even: enter while loop removing all instances.
    \item If odd: clear the list.
\end{enumerate}

The loop runs only if the count of \texttt{value} is even, and \texttt{value} exists in the list.
In the given test case, \texttt{`m'} appears 0 times (even), but is not in the list, so the loop does not execute.
To trigger execution, choose a character with even count $>0$.
From the string:
\begin{itemize}[topsep=0pt, itemsep=0pt]
    \item \texttt{`a'} appears 2 times
    \item \texttt{`b'} appears 4 times
    \item \texttt{`e'} appears 2 times
    \item \texttt{`u'} appears 2 times
    \item \texttt{`r'} appears 2 times
\end{itemize}

We can use any of these; choosing \texttt{`b'} since it appears 4 times.
\end{tcolorbox}
    
    \caption{Incorrect letter counting logic: Here, the model fails to correctly analyze string operations.}
    \label{fig:example-1}
\end{figure*}

%% file: diagrams/Example2.tex
\begin{figure*}
    \centering
\begin{tcolorbox}[colback=white, colframe=black, boxrule=0.8pt]
\begin{minipage}{0.6\textwidth}
\lstset{style=exstyle}
\begin{lstlisting}[language=Python,numbers=left,numberstyle=\tiny,stepnumber=1,numbersep=8pt]
import unittest

def f(text):
    created = []
    for line in text.splitlines():
        if line == '':
            break
        created.append(list(list(line.rstrip())[::-1][flush]))
    return created [::-1]

flush = 0

# Test Assertion
unittest.TestCase().assertEqual(f('A(hiccup)A'), [['A']])
\end{lstlisting}
\end{minipage}
\hfill
\begin{minipage}{0.35\textwidth}
\textbf{Program:} CRUXEval-712 \\[4pt]
\textbf{Uncovered Branch:} 6$\rightarrow$7 \\[4pt]
\textbf{Target:} Branch at line 6 \\[6pt]

\textbf{LLM Answer:}
\texttt{`A\textbackslash n'} \\[6pt]

\textbf{Error:} Native API Misevaluation
\end{minipage}

\end{tcolorbox}

\vspace{6pt}

\begin{tcolorbox}[colback=gray!10, colframe=black, title=\textbf{Chain-of-Thought Reasoning Trace of Claude Sonnet 4}]
Let me analyze the program step by step:
\begin{enumerate}[topsep=0pt, itemsep=0pt]
    \item The function \texttt{f(text)}` takes a text parameter.
    \item It splits the text into lines using \texttt{text.splitlines()}.
    \item It iterates through each line in a for loop (line 5).
    \item Line 6 is the conditional: \texttt{if line == `'}.
    \item If this condition is true (line is empty), it breaks out of the loop
    \item Otherwise, it processes the line and appends to \texttt{created}. To reach line 6, I need the function to be called with text that, when split into lines, produces at least one line to iterate over the for loop.
\end{enumerate}

Looking at the current test case: \texttt{f(`A(hiccup)A')}
\begin{itemize}[topsep=0pt, itemsep=0pt]
    \item \texttt{`A(hiccup)A'.splitlines()} returns \texttt{`A(hiccup)A'} (a list with one element)
    \item The for loop will iterate once with \texttt{line `A(hiccup)A'}
    \item Line 6 will be reached and will check \texttt{if `A(hiccup)A' == `':}
    \item This evaluates to \texttt{False}, so the \texttt{break} does not execute. To make the body of the conditional at line 6 execute (the \texttt{break} statement), I need a line that equals an empty string. This can happen if:
    \begin{itemize}[topsep=0pt, itemsep=0pt]
        \item The text contains an empty line (two consecutive newlines)
        \item The text starts with a newline
        \item The text ends with  a newline followed by nothing.         
    \end{itemize}
\end{itemize}

The simplest way is to provide text that contains an empty line. For example, \texttt{`A\textbackslash n'} would create \texttt{[`A', `']} when split; \texttt{`A\textbackslash n\textbackslash n'} would create \texttt{[`A', `', `']} when split; \texttt{`\textbackslash n'} would create \texttt{[`', `']} when split. Let me use \texttt{`A\textbackslash n'} which will create two lines \texttt{[`A', `']}. The loop will first process `A', then on the second iteration, \texttt{line} will be \texttt{`'}, making the condition \texttt{line == `'} true, thus executing the \texttt{break} statement.
\end{tcolorbox}
    
    \caption{Incorrect \texttt{splitlines()} analysis: Here, the model misunderstands the edge cases in text processing.}
    \label{fig:example-2}
\end{figure*}